\documentclass{ifacconf}

\usepackage{graphicx}      
\usepackage{subcaption}
\usepackage{float}
\usepackage{natbib}        
\usepackage{amsmath}
\usepackage{enumerate}
\usepackage{amssymb}

\begin{document}
\begin{frontmatter}

\title{Regularized GLISp for sensor-guided human-in-the-loop optimization\thanksref{footnoteinfo}} 

\thanks[footnoteinfo]{This work is partially supported by the FAIR project (NextGenerationEU, PNRR-PE-AI, M4C2, Investment 1.3), the 4DDS project (Italian Ministry of Enterprises and Made in Italy, grant F/310097/01-04/X56), and the PRIN PNRR project P2022NB77E (NextGenerationEU, CUP: D53D23016100001). It is also partly supported by the ENFIELD project (Horizon Europe, grant 101120657).}

\author[PoliMI]{Matteo Cercola}, 
\author[PoliMI]{Michele Lomuscio},
\author[IDSIA]{Dario Piga},
\author[PoliMI]{Simone Formentin}

\address[PoliMI]{Dipartimento di Elettronica, Informazione e Bioingegneria, Politecnico di Milano, Milano, Italy.(e-mail: author@polimi.it).}
\address[IDSIA]{IDSIA Dalle Molle Institute for Artificial Intelligence Research - SUPSI, Via la Santa 1, Lugano, 6900, Switzerland}

\begin{abstract}                
Human-in-the-loop calibration is often addressed via preference-based optimization, where algorithms learn from pairwise comparisons rather than explicit cost evaluations. While effective, methods such as Preferential Bayesian Optimization or Global optimization based on active preference learning with radial basis functions (GLISp) treat the system as a black box and ignore informative sensor measurements.
In this work, we introduce a sensor-guided regularized extension of GLISp that integrates measurable descriptors into the preference-learning loop through a physics-informed hypothesis function and a least-squares regularization term. This injects grey-box structure, combining subjective feedback with quantitative sensor information while preserving the flexibility of preference-based search. Numerical evaluations on an analytical benchmark and on a human-in-the-loop vehicle suspension tuning task show faster convergence and superior final solutions compared to baseline GLISp.
\end{abstract}


\end{frontmatter}

\section{Introduction}

Preference-based optimization provides a principled framework for addressing problems in which the decision-maker’s objective function is unknown and cannot be quantitatively  evaluated. 
Instead, the decision-maker expresses qualitative judgments in the form of pairwise comparisons, such as ``this option is preferred to that one''.  
This setting arises in many domains where performance is inherently subjective or difficult to quantify. 
Such preference‐based optimization has been applied in many areas, from calibrating control parameters in robotics or vehicle systems to tuning perceptual properties in visual prostheses, see, e.g., \cite{Tolasa2025HiLMultiModalHaptics,Wang2025RealtimePrefOpt,Granley2023HILVisualProstheses,catenaro2023active} and the references therein.

Several approaches have been proposed in the literature for preference-based optimization. The most popular methods belong to the class of Preferential Bayesian Optimization (PBO), which uses probabilistic surrogates (e.g., Gaussian processes) to model latent utilities and derive queries (see, e.g., \cite{Chu2005PreferenceGP,Brochu2010Interactive,Gonzalez2017PBO,benavoli2023bayesian}). Another promising approach is GLISp (Global optimization based on active preference learning with radial basis functions)~\cite{Bemporad2021GLISp}, which avoids probabilistic modeling in favor of simpler surrogate-based strategies combined with distance-based exploration terms. 

However, these methods  commonly treat the optimization problem as purely black-box, relying exclusively on user's preferences.  
Despite their success, they ignore auxiliary measurable quantities often available in practical systems. For example, in calibration of automotive systems one may have sensor data such as accelerations or angular rates, or, in physiological optimization settings, measurable biomarkers can be available. Ignoring this quantitative information may  lead to \textit{inefficiencies} in terms of number of queries required to reach the optimal solution.

Recent works have thus begun exploring ``grey‐box'' or ``model‐informed prior'' optimization, which incorporate prior models, soft constraints, or measurable signals to guide the search. For instance, \cite{Liao2025HOMI}  learns from synthetic users or prior knowledge to accelerate adaptation. Also, dual‐level Gaussian Process optimization methods (e.g. in geometry or molecular design) incorporate physically inspired priors into the mean functions or kernel structure, see \cite{Wang2023DualLevelGP}.

However, there remains a gap between preference-based methods (which focus on respecting human judgments) and grey-box/model-informed methods (which focus on physical or measurable descriptors). In many real-world tasks, both sources of information are available: subjective preferences and measurable sub-criteria or sensor signals that are correlated with subjective utility. 

Formally, let $f$ be an objective function to be minimized, which is not directly observable but depends on a set of measurable sub-objectives:
\begin{equation*}
    f = g(J_1, J_2, \ldots, J_p),
\end{equation*}
where $g$ denotes an aggregation mapping and $\{J_1, \ldots, J_p\}$ are sub-objectives, some of which can be estimated from data, simulations, or sensor measurements. Such a decomposition occurs naturally in many optimization problems: the global objective is unknown and inferred through preferences, yet it is often driven by a small number of physically meaningful descriptors. 
For example, in vehicle dynamics, perceived ride comfort is influenced by measurable signals such as RMS vertical acceleration or pitch rate. 
Similarly, in social media recommendation systems, user satisfaction is subjective but shaped by watch time, likes, and shares. 
These examples illustrate a common structure: although the global criterion is subjective, it is guided by a limited set of quantifiable subfunctions.

Integrating these two sources poses several methodological challenges: how to design the surrogate; how to induce a sensor-based bias in the preference model without overriding it; how to adaptively weight the influence of the physical prior when it is only partially reliable or even partially misleading. The methodological challenge is to infer the latent mapping between subfunctions and the global objective from preferences while leveraging the quantitative measurements as informative priors to guide the search. 
The framework must respect qualitative human judgments while incorporating the structure offered by measurable subfunctions. 
It should also adapt to situations where measurements  and preferences are only partially consistent, ensuring robustness rather than biasing the search toward misleading priors.

This motivates the development of a sensor-guided regularized extension of GLISp, which augments standard GLISp by introducing a hypothesis function based on measurable sub‐functions (sensor data), and a regularization term that penalizes disagreement between the surrogate learned from preferences and this hypothesis. An adaptive cross‐validation scheme tunes the strength of regularization to ensure robustness. To the best of our knowledge, this is the first preference-based scheme that injects measurable structure into GLISp via adaptive regularization, effectively transforming it from a black-box to a grey-box optimization framework.

We validate our method on both analytical optimization benchmarks and a practical vehicle suspension tuning task. The resulting framework achieves faster convergence, reduces user effort, and improves interpretability by revealing how measurable descriptors contribute to subjective preferences.

\section{GLISp: an overview}\label{sec:background}
In this section, we provide a short overview of the GLISp algorithm, originally proposed in  \cite{Bemporad2021GLISp}  for black-box preference-based optimization. 
Unlike preferential Bayesian Optimization, which models latent utility probabilistically, GLISp fits a radial-basis surrogate that satisfies a collected set of user's pairwise preferences  by solving a convex quadratic program. The surrogate is then used within an acquisition strategy that balances exploitation of the current model with exploration of unexplored regions. 

The GLISp framework consists of three main components:
\begin{enumerate}[i]
    \item \textbf{Preference model}, which encodes the decision-maker's qualitative judgments as follows. 
    Given two decision vectors $\mathbf{x}_A$ and $\mathbf{x}_B$, user feedback is encoded as:
    \begin{equation} \label{eq:pref_rule}
    \pi(\mathbf{x}_A, \mathbf{x}_B) =
    \begin{cases}
        -1 & \text{if } \mathbf{x}_A \text{ is preferred to } \mathbf{x}_B, \\
        0  & \text{if } \mathbf{x}_A \text{ and } \mathbf{x}_B \text{ are  equivalent}, \\
        +1 & \text{if } \mathbf{x}_B \text{ is preferred to } \mathbf{x}_A.
    \end{cases}
    \end{equation}
    Formally, assume that the user has collected a set $\mathcal{D} = \{\mathbf{x}_j\}_{j=1}^{N}$ of $N$ decision vectors, with $\mathbf{x}_j \in \mathcal{X} \subseteq \mathbb{R}^n$, and has evaluated a set of preferences $b_h \in \{-1,0,1\}$, with $h=1,\ldots,M \leq \binom{N}{2}$, according to the preference rule  \eqref{eq:pref_rule}, i.e.,
    \begin{equation}
    b_{h} = \pi(\mathbf{x}_{k(h)}, \mathbf{x}_{j(h)}),
    \label{eq:pref-vector}
    \end{equation}
    where $k(h), j(h) \in \{1,\ldots,N\}$ and $k(h) \neq j(h)$. 

    \item \textbf{Surrogate model}, which  approximates the latent utility function while satisfying collected preferences through an RBF approximation:
    \begin{equation}
    \hat{f}(\mathbf{x}) = \sum_{i=1}^{N} \beta_i \, \phi(\epsilon \, \| \mathbf{x} - \mathbf{x}_i \|_2^2),
    \end{equation}
    where $\{\mathbf{x}_i\}_{i=1}^N$ are the sampled points, $\phi(\cdot)$ is an RBF kernel~\cite{Gut01}, and $\epsilon>0$ is a kernel width. 

    Each comparison $\pi(\mathbf{x}_{i(h)}, \mathbf{x}_{j(h)})$ imposes the following constraints on the surrogate (with margin $\sigma > 0$ and slack variables $\xi_h \ge 0$):
    \begin{subequations} \label{eq:const}
        \begin{align}
    b_h = -1: & \quad \hat{f}(\mathbf{x}_{i(h)}) \leq \hat{f}(\mathbf{x}_{j(h)}) - \sigma + \xi_h,  \\
    b_h = +1: & \quad \hat{f}(\mathbf{x}_{j(h)}) \leq \hat{f}(\mathbf{x}_{i(h)}) - \sigma + \xi_h, \\
    b_h = 0: & \quad  \big|\hat{f}(\mathbf{x}_{i(h)}) - \hat{f}(\mathbf{x}_{j(h)})\big| \leq \sigma + \xi_h.
    \end{align}
    \end{subequations}

    The coefficients $\boldsymbol{\beta}=[\beta_1,\ldots,\beta_N]^T$ are obtained by solving:
    \begin{equation} \label{eqn:cost_ss}
    \min_{\boldsymbol{\beta}, \boldsymbol{\xi}} \quad  \sum_{h=1}^{M} \xi_h^2 + \lambda \sum_{k=1}^{N} \beta_k^2 \quad \text{s.t. constraints \eqref{eq:const}}, 
    \end{equation}
    where  $\lambda>0$  is a regularization parameter which guarantees that the cost function~\eqref{eqn:cost_ss} is strictly convex  and  admits a unique solution. Note that program \eqref{eqn:cost_ss} is a quadratic program, since the constraints \eqref{eq:const} are linear in the optimization variables $\boldsymbol{\beta}$ and $ \boldsymbol{\xi}$. 
    \item \textbf{Acquisition strategy}, which aims at selecting  the next query point $\mathbf{x}_{N+1}$ by minimizing:
    \begin{equation} \label{eq:acquisition}
    a(\mathbf{x}) = \frac{\hat{f}(\mathbf{x})}{\Delta \hat{f}} - \delta \, z(\mathbf{x}),
    \end{equation}
    where $\delta \ge 0$ controls exploration,  $z(\mathbf{x})$ is an inverse distance weighting (IDW) term  promoting sampling in unexplored regions and defined as
    \begin{equation}
z(\mathbf{x}) =
\begin{cases}
0 & \!\!\! \!\!\!\!\!\!  \mathbf{x} \in \{\mathbf{x}_1,\dots,\mathbf{x}_N\}, \\
\arctan \!\left( \dfrac{1}{\sum_{i=1}^N \| \mathbf{x} - \mathbf{x}_i\|_2^{-2}} \right) & \text{otherwise}, 
\end{cases}
\end{equation}
and  $\Delta \hat{f} \triangleq \max_{x \in \mathcal{X}} \hat{f}(x) - \min_{x \in \mathcal{X}} \hat{f}(x)$ denotes the range of the surrogate and serves as a normalization factor to facilitate the tuning of the exploration parameter  $\delta \in (0,1]$.
\end{enumerate}

Starting from an initial design (e.g., Latin Hypercube Sampling) with a set of $N$ decision vectors $\mathcal{D} = \{\mathbf{x}_j\}_{j=1}^{N}$ and related preferences expressed by the user, GLISp iteratively:
\begin{enumerate}
    \item Updates the surrogate by solving the quadratic program~\eqref{eqn:cost_ss},
    \item Selects a new candidate $\mathbf{x}_{N+1}$ by minimizing the acquisition function~\eqref{eq:acquisition},  
    \item Queries the user for a new comparison between candidate $\mathbf{x}_{N+1}$ and the current best solution.
\end{enumerate}
After the query budget is exhausted, GLISp outputs either the best sampled configuration or the minimizer of the final surrogate. In the following section,  we introduce a regularized extension of GLISp in order to exploit quantitative information affecting the utility function $f$.
\section{Sensor-Guided Preference Learning}\label{sec:method}

In this section, we present a \emph{sensor-guided, regularized} extension of the GLISp algorithm. 
The proposed method addresses the intrinsic limitation of purely black-box preference optimization by embedding measurable, physically interpretable descriptors into the optimization loop. 
Concretely, we augment the GLISp step of surrogate model construction  with a physics-informed regularization term that encourages agreement between the user's preferences  and a hypothesis function built from sensor-derived subobjectives. 
This modification transforms the problem from a purely black-box to a \emph{grey-box setting}: user preferences remain central, but they are systematically complemented by quantitative information that can accelerate learning, improve robustness, and enhance interpretability.

\begin{figure}[h!]
    \centering
    \includegraphics[width=0.556\textwidth,trim=155 0 60 40,clip]{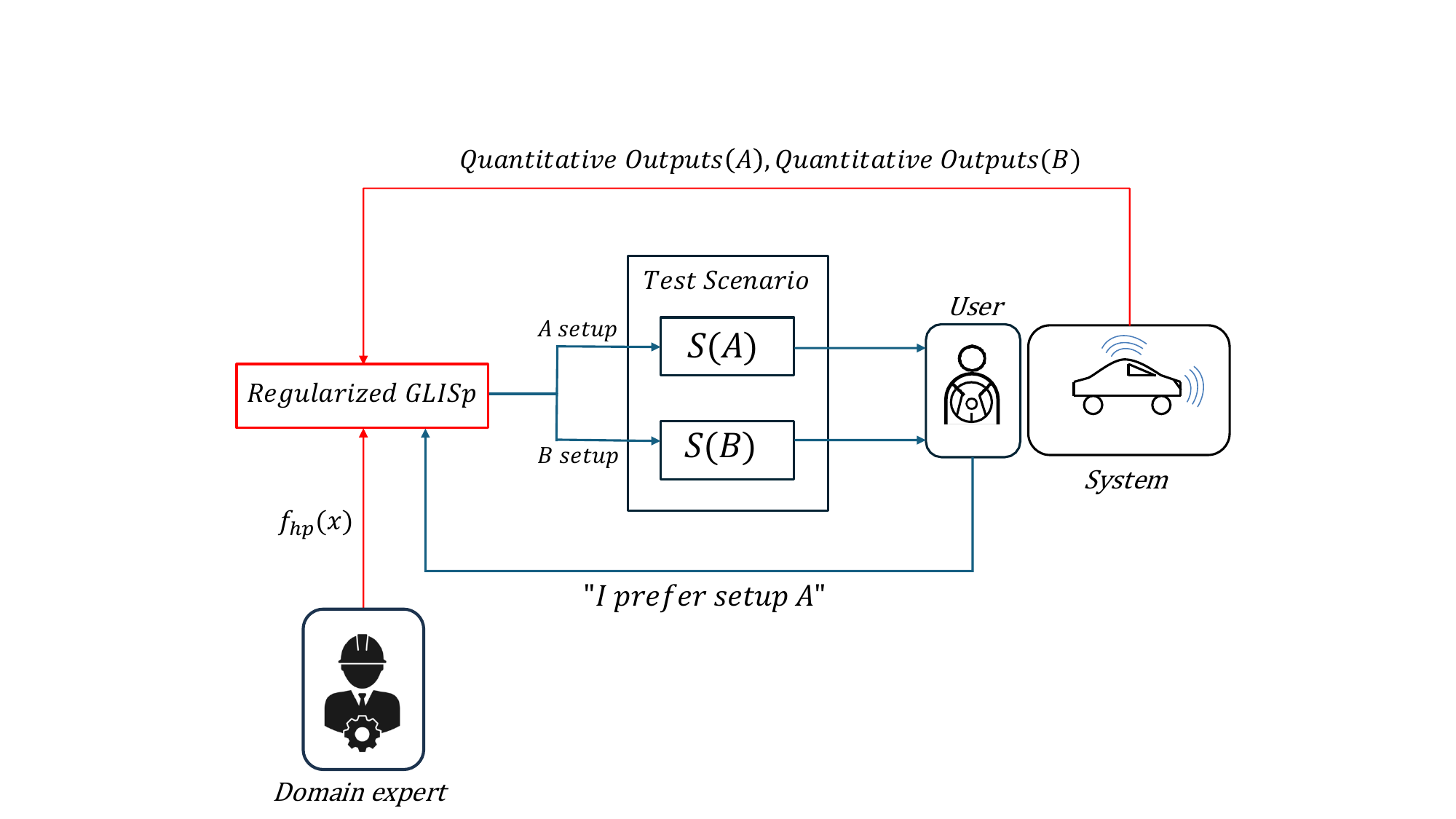}
    \caption{Sensor-guided regularized GLISp integrating quantitative sensor data into the preference-learning loop. User indicates the preferred setup, while domain expert defines the hypothesis function $f_{hp}$.  New components w.r.t. baseline GLISp are highlighted in red.}
    \label{fig:Regularize_framework}
\end{figure}

Fig.~\ref{fig:Regularize_framework} sketches the architecture of the proposed sensor-guided GLISp. 
The standard preference-based loop is preserved (surrogate update, acquisition strategy, query), while an additional information channel (computed from measurable descriptors) enters the surrogate fitting stage and influences the learned model through regularization.

\subsection{Hypothesis function}
We introduce a \emph{hypothesis function} $f_{hp}:\mathcal{X}\to\mathbb{R}$ that maps decision vectors to a quantitative estimate of the latent utility function $f$ based on $p$ measurable subfunctions:
\begin{equation}
\label{eq:hp_linear}
f_{hp}(\mathbf{x}) = \sum_{r=1}^{p} w_r \, J_r(\mathbf{x}),
\end{equation}
where $J_r(\mathbf{x})$ are descriptors computable from sensor data, simulations, or domain models, and $\mathbf{w}=[w_1,\dots,w_p]^\top$ are weights that encode the relative importance of each descriptor.

Modeling $f_{hp}$ as a parametric combination of descriptors has three practical advantages. 
First, it injects domain knowledge into the optimization loop in a transparent way: each $w_r$ quantifies the contribution of descriptor $J_r$. 
Second, it reduces sample complexity when descriptors are informative, because the hypothesis already encodes useful structure. 
Third, learning $\mathbf{w}$ jointly with the surrogate enables the algorithm to adapt the prior online: if a descriptor is irrelevant or misleading, its learned weight will be reduced.

While we focus on the linear form \eqref{eq:hp_linear} for interpretability and tractability, the framework admits richer hypothesis classes (e.g., generalized linear models, basis expansions, or neural-network parametrizations). 

\subsection{Physics-informed regularization}
To exploit the hypothesis $f_{hp}$ within GLISp, we add a regularization term to the surrogate fitting objective that penalizes disagreement between the RBF surrogate $\hat f$  and $f_{hp}$ evaluated at the sampled points $\{\mathbf{x}_i\}_{i=1}^N$. The joint fitting problem becomes:
\begin{equation}
\label{eq:reg_obj_repeat}
\begin{aligned}
\min_{\boldsymbol{\beta},\boldsymbol{\xi},\mathbf{w}}\quad
& \sum_{h=1}^{M}\xi_h^2 \;+\; \lambda_{\beta}\sum_{k=1}^{N}\beta_k^2
\;+\; \lambda_{LS}\sum_{i=1}^{N}\big(\hat f(\mathbf{x}_i)-f_{hp}(\mathbf{x}_i)\big)^2 \\
\text{s.t.}\quad
& \text{preference constraints } \eqref{eq:const}, 
\end{aligned}
\end{equation}
 where $\lambda_{\beta},\lambda_{LS}\ge 0$ are regularization hyperparameters controlling  slack penalization and the strength of the physics-informed alignment, respectively.
 A few remarks are in order:
\begin{itemize}
   \item \textbf{Preservation of preferences.} The preference constraints remain active and continue to be enforced as in the original GLISp formulation: their feasibility is preserved up to the admissible slack variables $\xi_h$. The hyperparameter $\lambda_{LS}$ regulates the trade-off between adherence to the physics-informed hypothesis and fidelity to user preferences: when $\lambda_{LS}=0$ the method reduces exactly to standard GLISp, while large values of $\lambda_{LS}$ bias the surrogate strongly toward the hypothesis $f_{hp}$. However, in this latter case, if the hypothesis is only partially consistent with the actual preferences, the optimizer may choose to satisfy the regularization term at the cost of activating slack variables, thus risking a surrogate $\hat{f}$ that does not fully comply with some preference constraints. This motivates the need for an adaptive or data-driven tuning of $\lambda_{LS}$, which will be discussed later.
    \item \textbf{Joint estimation.} The optimization variables include both surrogate coefficients $\boldsymbol{\beta}$ and hypothesis weights $\mathbf{w}$. Joint estimation allows the algorithm to adapt the physical prior as more preference information becomes available.
    \item \textbf{Computational considerations.} When $f_{hp}$ is linear in $\mathbf{w}$ (as in \eqref{eq:hp_linear}), the regularization term is quadratic in $(\boldsymbol{\beta},\mathbf{w})$ and thus the learning  problem \eqref{eq:reg_obj_repeat} remains convex.
\end{itemize}

\subsection{Hyperparameter tuning via cross-validation} \label{sec:cross-validation}
Selecting a fixed $\lambda_{LS}$ a priori is generally suboptimal because the informativeness of descriptors may vary not only across tasks but also dynamically during the iterations of the regularized GLISp algorithm. For this reason, we adopt an adaptive $K$-fold cross-validation (CV) scheme that periodically re-estimates $\lambda_{LS}$ from the available preference data. The same validation mechanism is also used to tune the remaining hyperparameters of the surrogate (namely, the RBF width $\epsilon$ and the regularization $\lambda_\beta$ on the coefficients $\boldsymbol{\beta}$) ensuring a coherent data-driven calibration of the entire model.

Concretely, at predetermined intervals (e.g., every $T_{cv}$ iterations), we:
\begin{enumerate}
    \item form $K$ folds of the collected preference comparisons;
    \item for each candidate triplet $(\lambda_{LS}, \lambda, \epsilon)$ in a search grid, train the surrogate and hypothesis weights on $K-1$ folds and evaluate how many preference constraints in the held-out fold are satisfied by the learned surrogate;
    \item select the hyperparameter configuration that minimizes the average number of violated preferences across folds, and use it for the subsequent optimization iterations.
\end{enumerate}

By using the number of satisfied preferences on the validation fold as the selection metric, the procedure explicitly promotes models that generalize human judgments beyond the training comparisons. In this sense, regularized GLISp is expected to exhibit improved preference generalization compared to the standard (non-regularized) GLISp, especially when descriptors provide complementary structure that helps the surrogate extrapolate user preferences more reliably. This improves robustness especially  in early-stage learning when the number of preferences data is low.


The next section presents empirical evidence of these benefits on analytical and suspension-tuning benchmarks.
\section{Experimental Results}\label{sec:experiments}

This section presents a numerical evaluation of the proposed sensor-guided regularized GLISp on two complementary case studies. The first is a nonlinear analytical optimization problem, while the second is a realistic engineering scenario involving a human-in-the-loop vehicle suspension calibration task, where subjective ride comfort is the objective to maximize.

In all experiments, the regularized GLISp is compared against the baseline black-box version. A Monte Carlo study is performed, and for each run and at every iteration of the algorithm, we compute the optimization error defined as the difference between the best achieved value $y_{\rm best}$ and the global optimum $y^*$. The value of $y^*$ is obtained offline using numerical procedures requiring a large number of utility function evaluations (particle swarm optimization in the analytical problem and exhaustive grid search in the vehicle suspension calibration task). Results are reported in terms of mean and standard deviation over the Monte Carlo runs. 

The maximum number of iterations is fixed at each Monte Carlo run. For a fair comparison,   regularized and baseline GLISp share   the same initial design and maximum number of iterations. All the hyper-parameters of the GLISp algorithm are computed through the cross-validation approach described in Section~\ref{sec:cross-validation}.

\subsection{Analytical problem}

In the first case study, we minimize the following   seven-dimensional  ground-truth objective function:
\begin{equation}
\label{eq:analytical_function}
\resizebox{\columnwidth}{!}{$
f(\mathbf{x}) = \eta_0 x_0^4 - \eta_1(x_1-x_2)^2 + \dfrac{\eta_2}{x_3}
+ \eta_3\sin(x_4+x_5) + \eta_4 x_5 x_6,
$}
\end{equation}
where $\mathbf{x}=[x_0,\dots,x_6]^\top\in\mathbb{R}^7$ and $\mathbf{\eta}=[\eta_0,\dots,\eta_4]^\top$ are the ground-truth parameters. 
To test robustness and generality, the coefficients $\mathbf{\eta}$ are randomized at  each Monte Carlo run, producing a new optimization landscape every trial.

The hypothesis function $f_{hp}$ used by the regularized method adopts the same analytic form as $f$, with its own parameter vector $\mathbf{w}$ learned by solving the QP problem \eqref{eq:reg_obj_repeat}.  

Figure~\ref{fig:toy_results_agg} shows the aggregated convergence behaviour over 10 independent Monte Carlo runs. The regularized GLISp exhibits a significantly faster reduction of the optimization error compared to baseline GLISp. After 8 iterations, the mean error achieved by the regularized variant is approximately $0.45$, whereas the baseline reaches around $2.82$ at the same iteration. At the last iteration, the final error settles at about $0.28$ for the regularized method vs $1.90$ for the baseline. 

The shaded bands ($\pm$ standard deviation) are also  narrower for the sensor-guided version. In the final iteration, the standard deviation of baseline GLISp is approximately 2.5 times higher than that of the regularized variant, indicating more consistent performance of regularized GLISp across random landscapes. These results confirm that the physics-informed regularization not only accelerates convergence but also improves robustness by reducing variability.

\subsection{Vehicle suspension tuning}

To demonstrate applicability to a realistic engineering problem, we consider the problem of tuning suspension parameters of a half-car model with the objective of improving perceived ride comfort during a bump test. Comfort is quantified using two measurable features known to influence perception: the RMS of vertical acceleration $A_z$ and the RMS of pitch rate $\dot\theta$. 
The ground-truth objective is defined as a weighted combination of these features:
\begin{equation}
\label{eq:ground_truthCar}
f(\mathbf{x}) = \eta_{A_z}\sqrt{\frac{1}{T}\int_{0}^{T}A_z(t)^2\,dt}
+ \eta_{\dot\theta}\sqrt{\frac{1}{T}\int_{0}^{T}\dot\theta(t)^2\,dt},
\end{equation}
where $T$ is the simulation duration and $\eta_{A_z},\eta_{\dot\theta}>0$ set the relative importance of each term, which take different values at each Monte Carlo run. 
The hypothesis $f_{hp}$ used by the regularized method shares the  same functional form of Eq.~\eqref{eq:ground_truthCar}, with weighting parameters learned by solving problem \eqref{eq:reg_obj_repeat}. 

All simulations employ a half-car model implemented in MATLAB/Simulink and a repeatable bump input at $30$~km/h.  
To emulate human responses, a \emph{synthetic user} supplies pairwise preferences derived from the noise-free ground truth \eqref{eq:ground_truthCar}.  
This synthetic evaluator allows precise measurement of optimization performance while preserving the human-in-the-loop protocol of preference queries.

\subsubsection{2D suspension calibration.}
The 2D scenario involves tuning two damping coefficients: the front $c_f$ and the rear $c_r$. Figure~\ref{fig:2d_signals_full} shows the vehicle responses (vertical acceleration and pitch rate) obtained at the final iteration of both baseline and regularized GLISp, averaged over 10 Monte Carlo runs. We can observe that the configuration returned by the regularized GLISp yields a more damped and consistent transient immediately after the bump, whereas the baseline solution exhibits higher variability and slower settling. Notably, the vertical acceleration under the baseline configuration does not return to a zero steady state, indicating persistent oscillations. In practical terms, such residual vertical motion would translate into a discomfort effect for the driver, typically perceived as motion-induced nausea. 

\begin{figure*}[!t]
    \centering
    \begin{subfigure}[b]{0.32\textwidth}
        \centering
        \includegraphics[width=\linewidth]{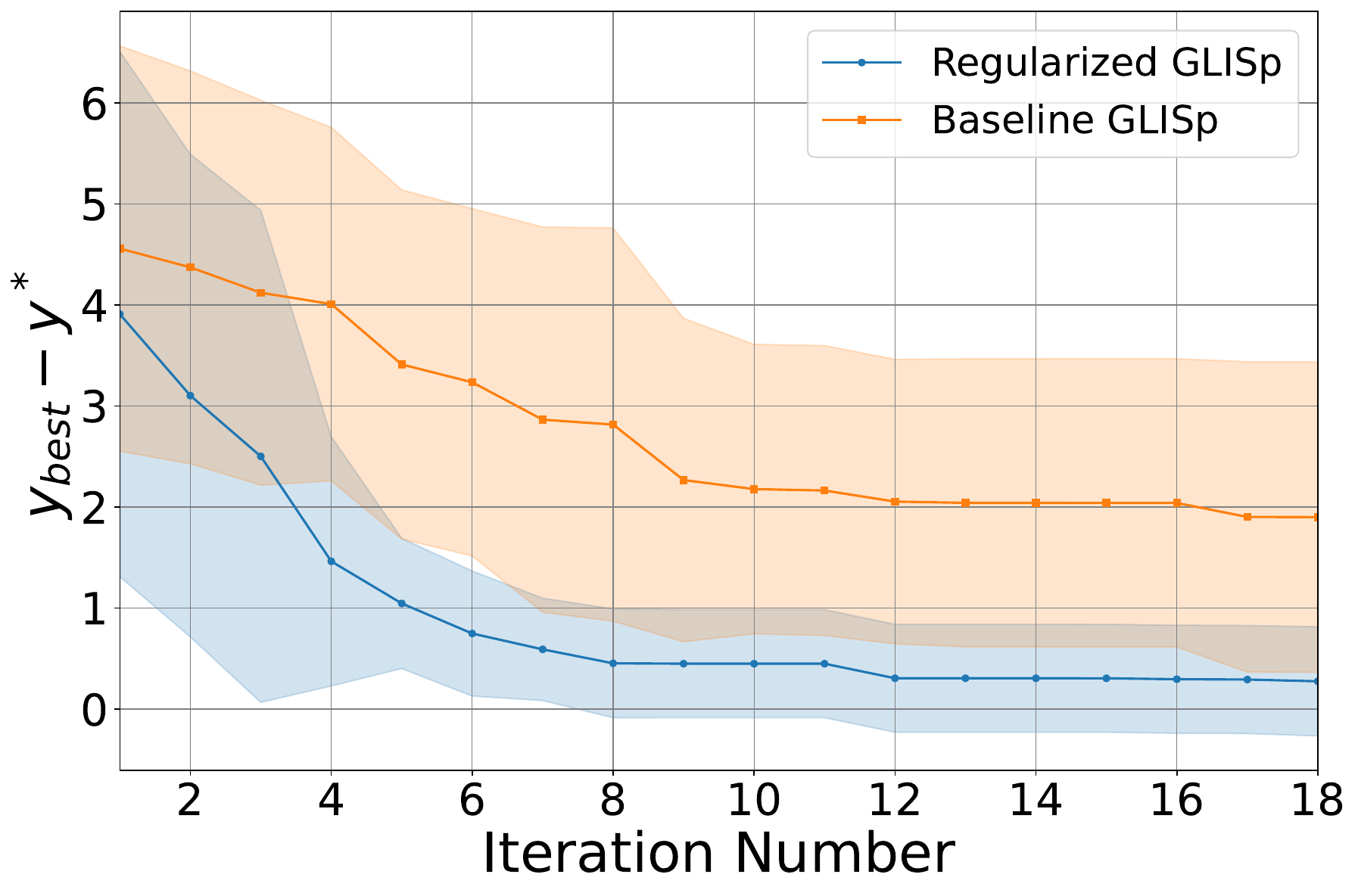}
        \caption{Analytical problem.}
        \label{fig:toy_results_agg}
    \end{subfigure}
    \hfill
    \begin{subfigure}[b]{0.32\textwidth}
        \centering
        \includegraphics[width=\linewidth]{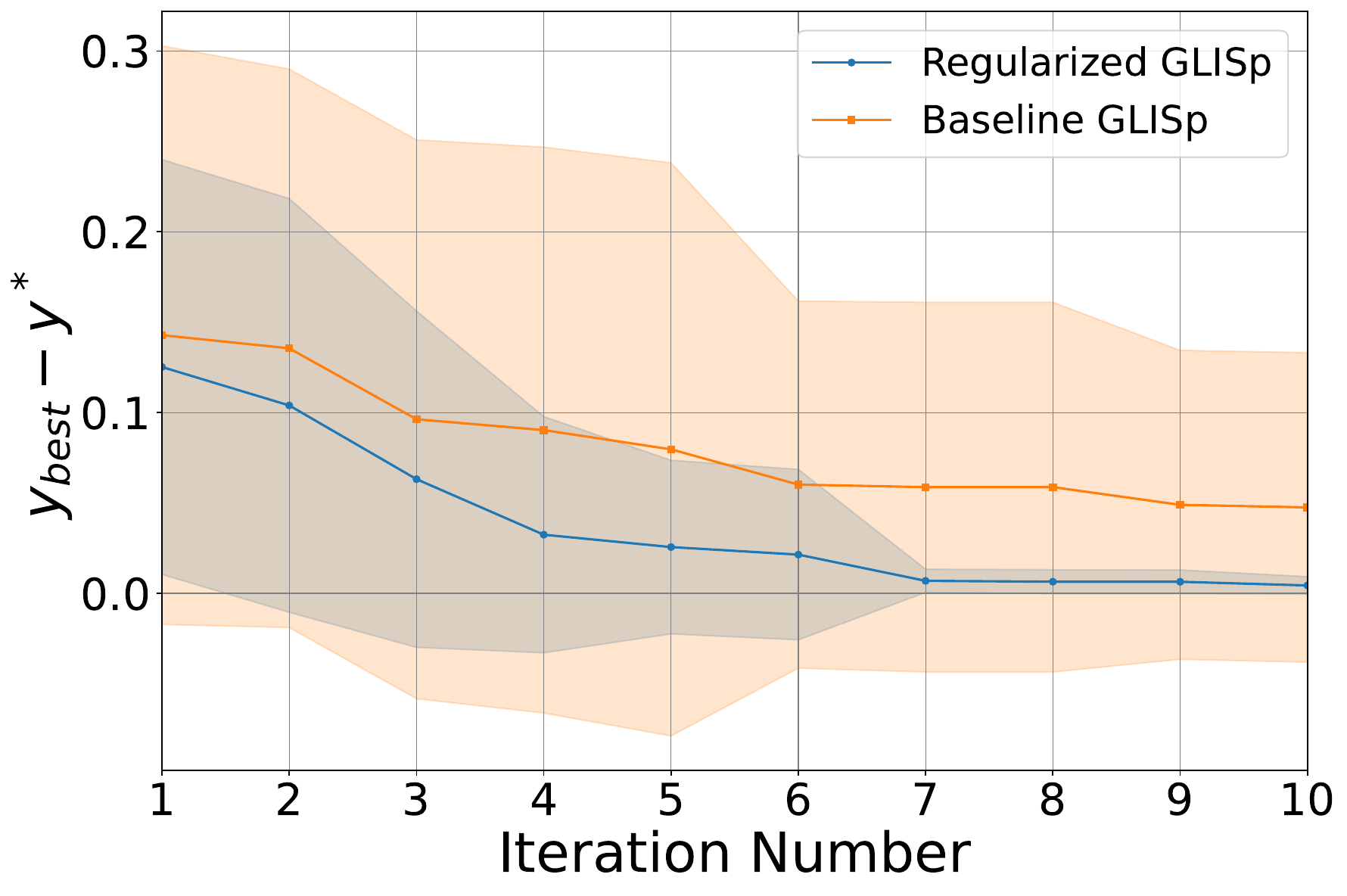}
        \caption{2D suspension tuning.}
        \label{fig:2d_results_agg}
    \end{subfigure}
    \hfill
    \begin{subfigure}[b]{0.32\textwidth}
        \centering
        \includegraphics[width=\linewidth]{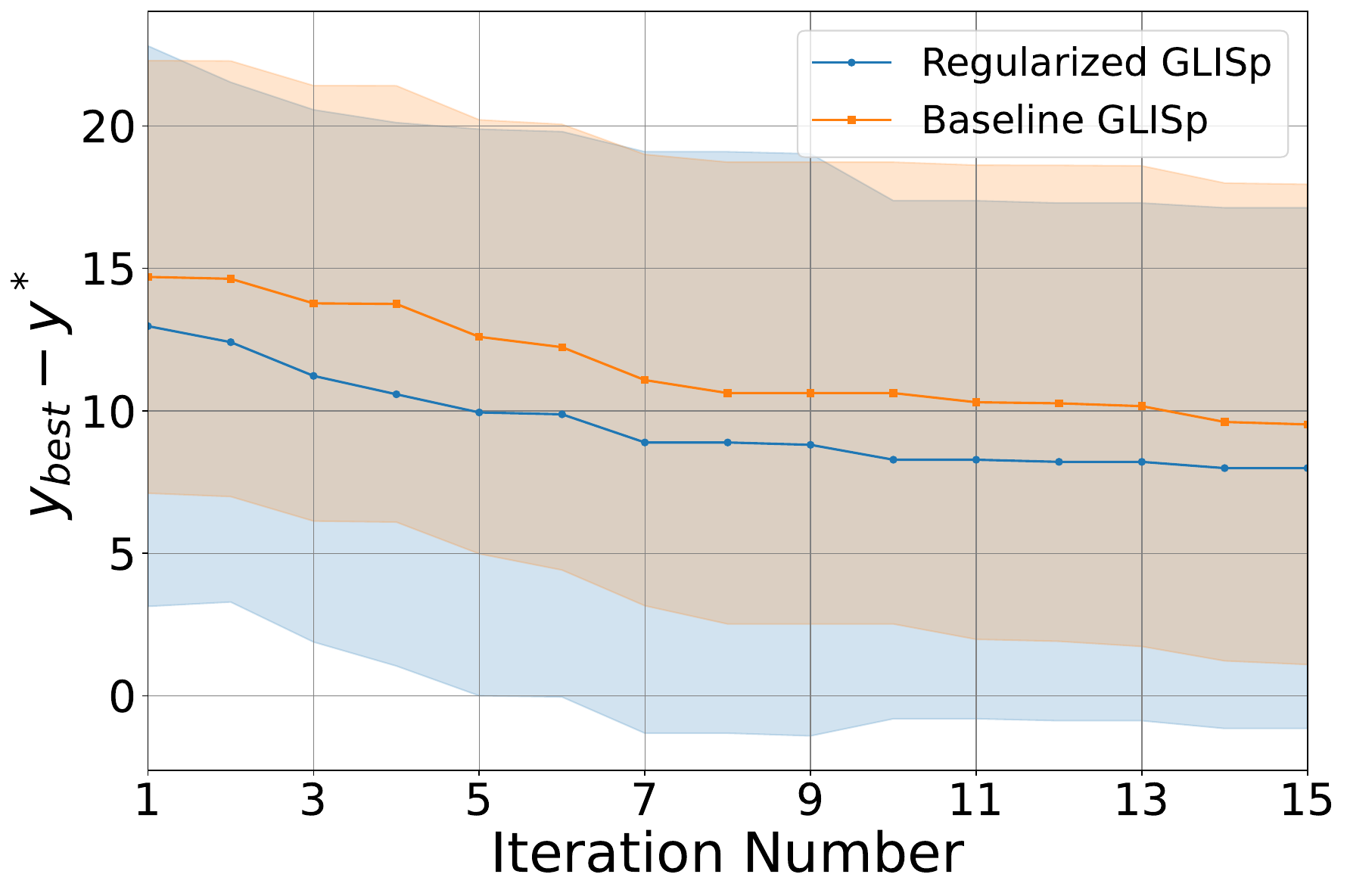}
        \caption{4D suspension tuning.}
        \label{fig:4d_results_agg}
    \end{subfigure}
   \caption{Difference between the best achieved value $y_{\rm best}$ at each iteration and  optimal value $y^*$:  mean (solid line) and $\pm$  standard deviation (shaded region) over 10 Monte Carlo simulations. Baseline GLISp (orange) vs. regularized GLISp (blue).}
    \label{fig:all_convergence_results}
\end{figure*}

Convergence statistics in Figure~\ref{fig:2d_results_agg} support this outcome. Indeed,  the sensor-guided formulation achieves a substantially lower optimization error with markedly fewer queries, reaching near-zero error in approximately 7 iterations on average. In contrast, the baseline  GLISp converges more slowly and frequently stagnates at higher error levels. At convergence, the final mean error is about 10 times lower than that of the baseline GLISp, and the variability is reduced by a factor of 17, confirming that the physics-informed regularization yields both faster and more stable convergence.


\subsubsection{4D suspension calibration.}
In the 4D scenario, the tuning problem is extended  to four parameters: the front and rear damping coefficients ($c_f$, $c_r$) and the front and rear stiffness ratios ($k_f$, $k_r$).
This combined stiffness–damping search space can yield configurations that compromise tire contact, potentially endangering vehicle safety.
For this reason, the ground-truth objective used in the 2D case study (Eq.~\eqref{eq:ground_truthCar}) is augmented with an additional penalty term accounting for \emph{grip loss}:

\begin{equation}
\label{eq:ground_truthCar_4D}
f(\mathbf{x}) =
\eta_{A_z}\sqrt{\frac{1}{T}\int_{0}^{T}A_z^2(t)\,dt}
+ \eta_{\dot{\theta}}\sqrt{\frac{1}{T}\int_{0}^{T}\dot{\theta}^2(t)\,dt}
+ \eta_{\text{grip}} \, T_{\text{loss}},
\end{equation}
where the first two terms coincide with those in Eq.~\eqref{eq:ground_truthCar}, and $T_{\text{loss}}$ denotes the total \emph{loss-of-grip time}, defined as the cumulative duration during which the tire force becomes non-positive.
The weighting coefficient $\eta_{\text{grip}}$ penalizes unsafe configurations exhibiting tire lift-off, ensuring that solutions violating the grip constraint are disfavored during optimization.
The hypothesis model $f_{hp}$, in contrast, encodes only comfort-related metrics and remains deliberately unaware of grip loss. Consequently, it is identical to that employed in the 2D suspension tuning case study.
This choice reflects a realistic limitation: sensors capable of detecting tire lift-off are typically expensive and seldom available.
As a result, the human evaluator acts as a safety supervisor, rejecting unsafe configurations through preference feedback.

\begin{figure*}[t]
    \centering
    \begin{subfigure}[b]{0.48\textwidth}
        \centering
        \includegraphics[width=\linewidth]{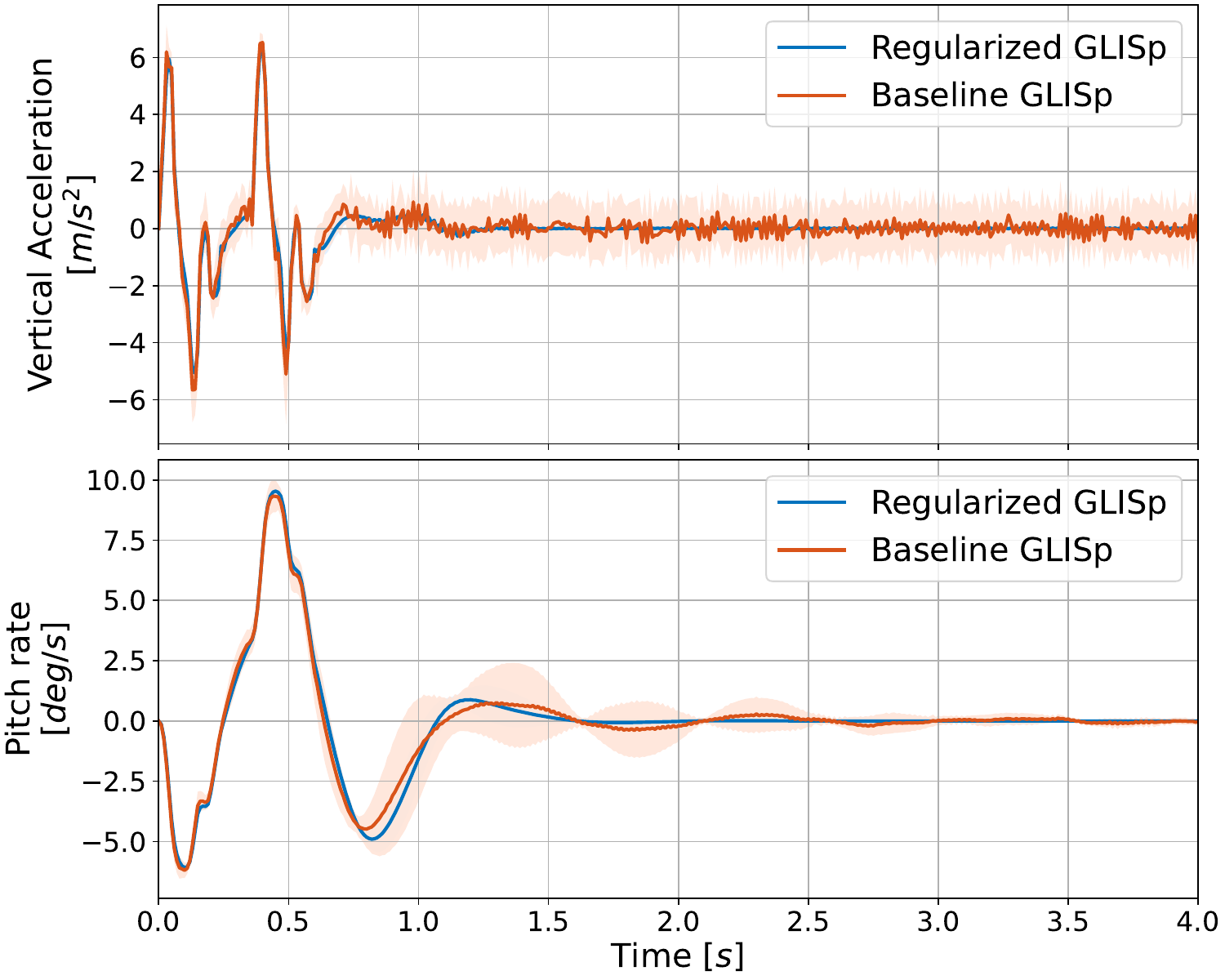}
        \caption{2D suspension calibration.}
        \label{fig:2d_signals_full}
    \end{subfigure}
    \hfill
    \begin{subfigure}[b]{0.48\textwidth}
        \centering
        \includegraphics[width=\linewidth]{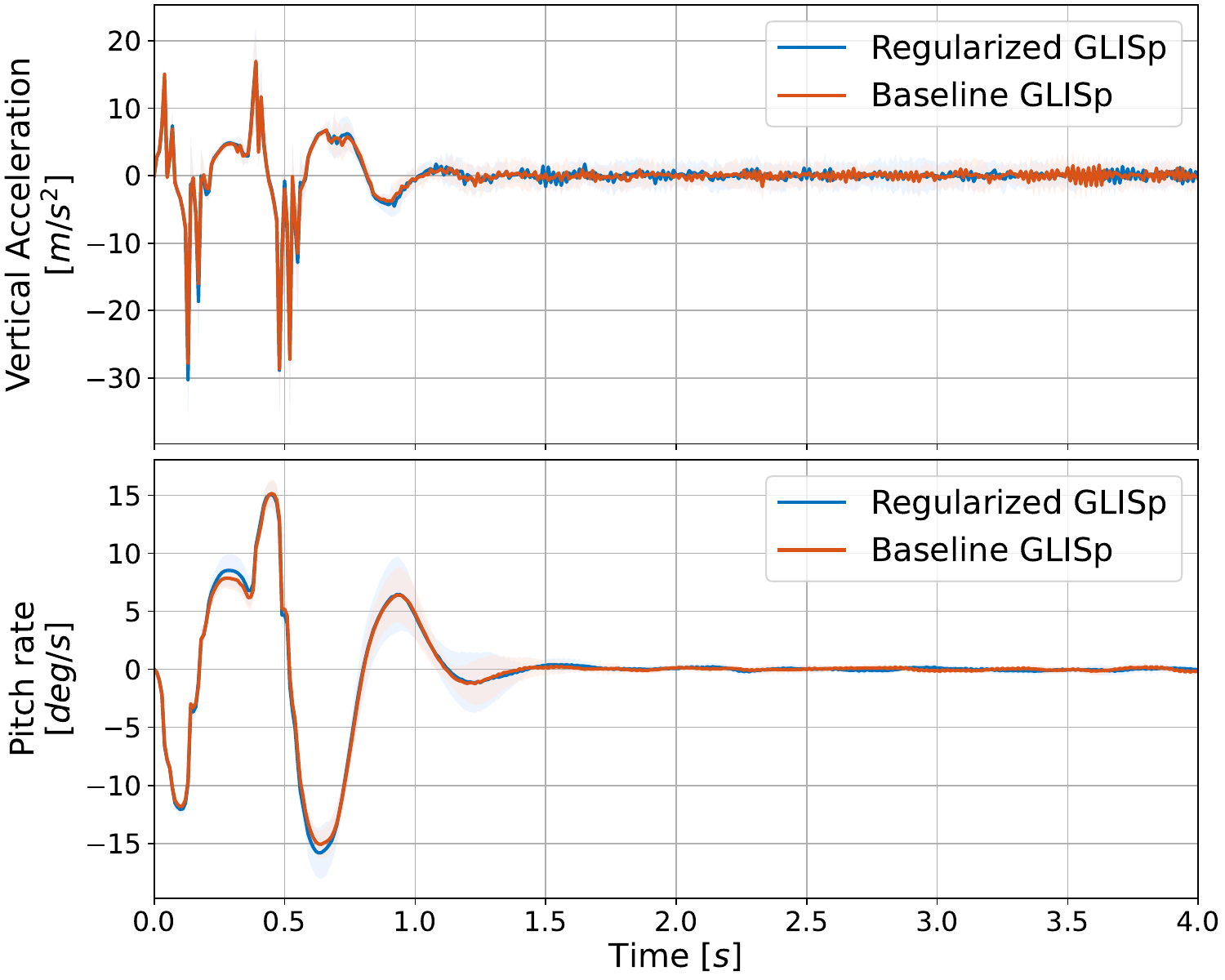}
        \caption{4D suspension calibration.}
        \label{fig:4d_signals_full}
    \end{subfigure}
       \caption{Vehicle response in terms of vertical acceleration (top) and pitch rate (bottom) obtained using the parameters learned at the final iteration of the baseline (orange) and regularized (blue) GLISp. Solid lines indicate the mean response, and the shaded bands represent $\pm$  standard deviation over 10 Monte Carlo runs.}
    \label{fig:2d_4d_fullwidth}
\end{figure*}

Figure~\ref{fig:4d_signals_full} reports the vehicle responses (vertical acceleration and pitch rate) obtained at the final iteration of both the baseline and the regularized GLISp, averaged over 10 Monte Carlo runs. The configuration returned by the regularized GLISp exhibits a faster settling response and smaller residual vertical motion.

Convergence statistics in Figure~\ref{fig:4d_results_agg} confirm this trend: the sensor-guided formulation achieves a lower mean error and reduced variance compared to the baseline GLISp, demonstrating superior performance in a higher-dimensional and more safety-critical search space. 
Overall, the obtained numerical results highlight the advantages of exploiting physically informative descriptors within a preference-based optimization loop.


\section{Conclusions}\label{sec:conclusion}

This work is the first contribution that systematically integrates sensor-derived structure into GLISp, thus shifting it from a purely black-box formulation to a grey-box paradigm and demonstrating that preference-based optimization can leverage physical measurements without overriding subjective feedback. Numerical evaluations on an analytical benchmark and on a human-in-the-loop vehicle suspension tuning task showed that the proposed approach accelerates convergence and improves robustness and consistency in preference learning, especially in early stages when only few comparisons are available.

One limitation of direct regularization is that enforcing pointwise agreement between surrogate and hypothesis functions may be overly restrictive, especially when the hypothesis is underparametrized and does not capture all relevant  variables that might influence user's preference. A promising research direction is therefore to move from value-based to variation-based regularization, where the surrogate is not forced to match the hypothesis, but only to preserve consistent sensitivity with respect to measurable descriptors. In this view, gradient-alignment penalties may act as softer structural constraints, allowing the preference model to exploit incomplete physical information without being biased by its limitations. Adapting the strength of this structural prior locally across the decision space could further improve robustness when sensor-derived indicators are only partially reliable, paving the way for more expressive forms of grey-box preference learning.

\bibliography{ifacconf}

\end{document}